\documentclass[10pt,twocolumn,letterpaper]{article}

\usepackage{wacv}
\usepackage[accsupp]{axessibility}
\usepackage{times}
\usepackage{epsfig}
\usepackage{graphicx}
\usepackage{amsmath}
\usepackage{amssymb}
\usepackage{booktabs}
\usepackage{algorithm}
\usepackage{algpseudocode}
\usepackage{float}
\usepackage{color}
\usepackage{colortbl}
\usepackage{pifont}
\usepackage{breqn}
\usepackage{rotating}
\usepackage{enumitem}
\usepackage{makecell}
\usepackage{lineno}
\usepackage{xcolor}
\usepackage{multirow}
\usepackage{hhline}
\usepackage{amsfonts}
\usepackage{empheq}
\usepackage{framed, color}
\usepackage{xcolor}
\usepackage{graphics}
\usepackage{graphicx}
\usepackage[]{graphicx} 
\usepackage{epsfig} 
\usepackage{filecontents}
\usepackage{verbatim} 
\usepackage{tabularx}
\usepackage{setspace}
\usepackage{textcomp,booktabs}
\usepackage{caption}
\usepackage{stmaryrd}
\usepackage[mathscr]{eucal}
\usepackage{bbm}
\usepackage[export]{adjustbox}

\def\wacvPaperID{826} 

\wacvalgorithmstrack   

\wacvfinalcopy 

\ifwacvfinal
\usepackage[breaklinks=true,bookmarks=false]{hyperref}
\else
\usepackage[pagebackref=true,breaklinks=true,colorlinks,bookmarks=false]{hyperref}
\fi

\def\D{{\bf D}}

\def\F{{\bf F}}

\def\K{{\bf K}}

\def\I{{\bf I}}

\def\T{{\bf T}}
\def\X{{\bf X}}

\def\P{{\bf P}}
\def\q{{\bf q}}

\def\x{{\bf x}}

\def\M{{\bf M}}

\def\W{{\bf W}}

\def\0{{\bf 0}}
\def\1{{\bf 1}}

\newcommand{\xmark}{\ding{55}}%

\def\argmin{\mathop{\rm argmin}}

\begin{document}

\title{GeoFill: Reference-Based Image Inpainting with Better Geometric Understanding}

\author{%
Yunhan Zhao\textsuperscript{1}\footnotemark, Connelly Barnes\textsuperscript{2}, Yuqian Zhou\textsuperscript{2,3}, Eli Shechtman\textsuperscript{2}, Sohrab Amirghodsi\textsuperscript{2}, Charless Fowlkes\textsuperscript{1} \\
$^1$UC Irvine  \qquad $^2$Adobe Research \qquad $^3$IFP, UIUC \\
{\tt\small \{yunhaz5, fowlkes\}@ics.uci.edu } \ {\tt\small \{cobarnes, elishe, tamirgho\}@adobe.com \ yuqian2@illinois.edu}
}

\maketitle
\thispagestyle{empty}

\renewcommand{\thefootnote}{\fnsymbol{footnote}}
\setcounter{footnote}{1} 
\footnotetext{Work done while an intern at Adobe.}
\setcounter{footnote}{0} 
\renewcommand*{\thefootnote}{\arabic{footnote}}

\begin{abstract}
    Reference-guided image inpainting restores image pixels by leveraging the content from another single reference image. The primary challenge is how to precisely place the pixels from the reference image into the hole region. 
    Therefore, understanding the 3D geometry that relates pixels between two views is a crucial step towards building a better model.
    Given the complexity of handling various types of reference images, we focus on the scenario where the images are captured by freely moving the same camera around. 
    Compared to the previous work, we propose a principled approach that does not make heuristic assumptions about the planarity of the scene.
    We leverage a monocular depth estimate and predict relative pose between cameras, then align the reference image to the target by a differentiable 3D reprojection and a joint optimization of relative pose and depth map scale and offset.
    Our approach achieves state-of-the-art performance on both RealEstate10K and MannequinChallenge dataset with large baselines, complex geometry and extreme camera motions. We experimentally verify our approach is also better at handling large holes. 
\end{abstract}

\section{Introduction}
\label{sec:intro}
Image inpainting aims to plausibly restore missing pixels within a given hole region. Existing single image inpainting models~\cite{yu2018generative,xiong2019foreground,yu2019free} solve this problem without additional information by leveraging the knowledge learned from large scale training data or existing patches within the image. These methods become less reliable when input images contains large holes and the filling regions are complex in structures and textures. 

In 2021, Zhou \textit{et al.} \cite{zhou2021transfill} proposed a novel inpainting task called reference-based image inpainting. It aims at filling the hole regions of a ``target" image using another single source photo taken of the same scene. This foreground removal application is particularly useful when people take photos at museums or famous landmarks where the background is unique. It is nearly impossible for existing single image inpainting to faithfully restore what was \emph{actually there in the background} for such cases. Inpainting with a reference image is appealing and indispensable, and also feasible, since other photos taken at different viewpoints of the same scene can be available, e.g. in the users' albums or even downloaded from the Internet.




\begin{figure} [t]
\centering
\includegraphics[width=0.99\linewidth]{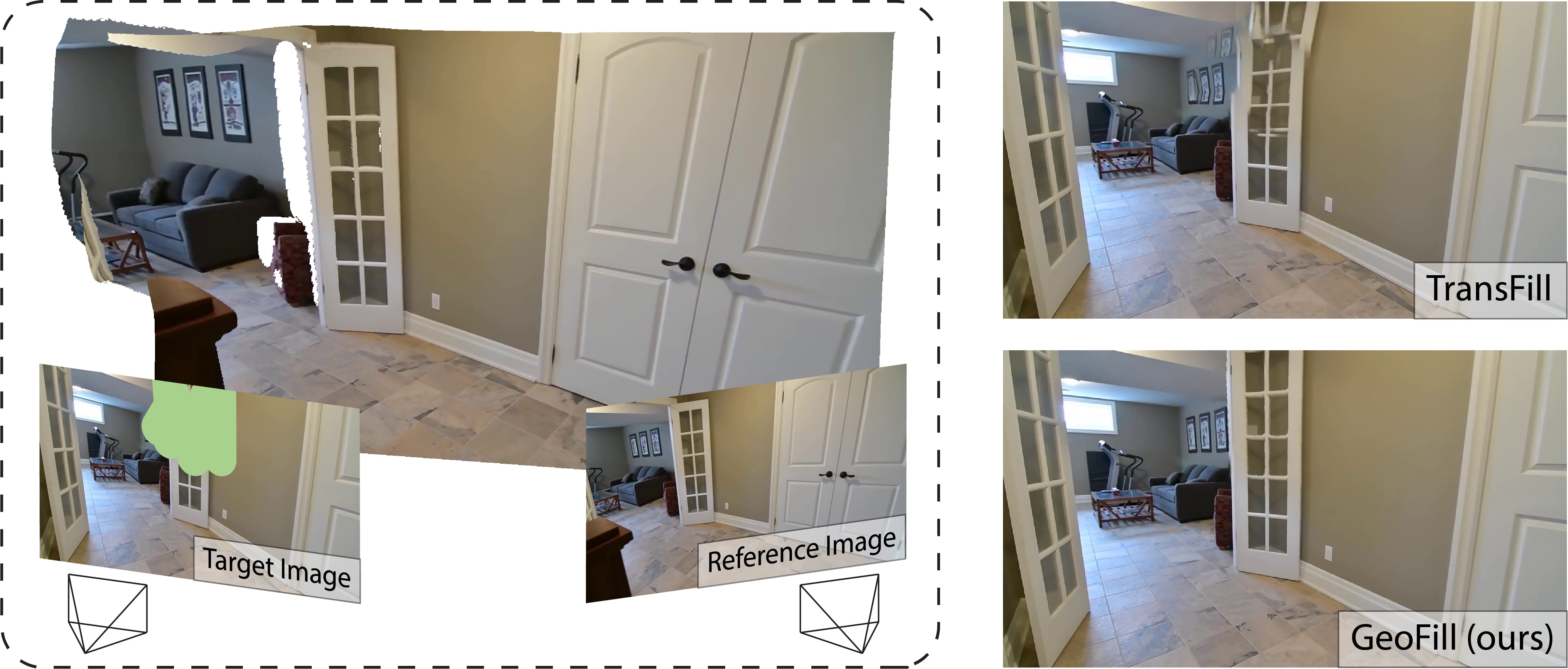}
\caption{
Given a reference image and a target image with hole, GeoFill utilizes predicted correspondence matching and depth maps to estimate a 3D mesh and relative camera pose and intrinsics. 
Compared to TransFill, the previous state-of-the-art approach, GeoFill handles complex scenes better by iteratively refining predicted depth maps and relative pose.  
}
\label{fig:teaser}
\vspace{-4mm}
\end{figure}

\begin{figure*}[t]
\centering

\includegraphics[width=0.975\textwidth]{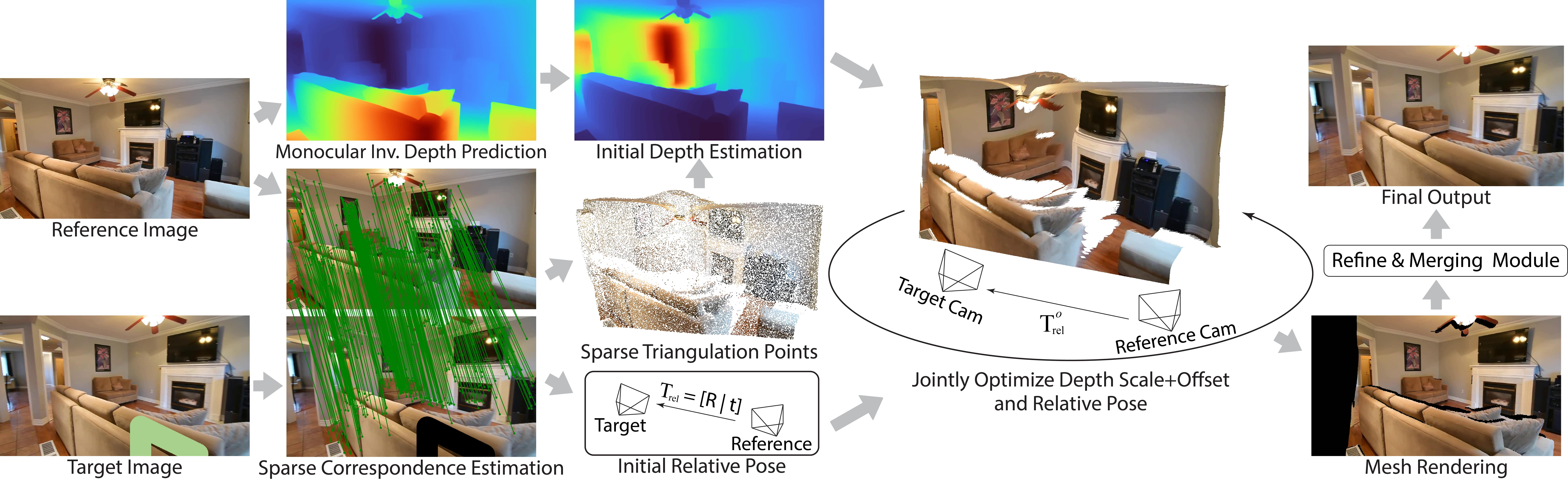}
\caption{
Overview of our system pipeline. We are given a target image with a hole and a source image. We aim at warping the single-source image to the target to fill the hole. We first estimate the relative pose as well as predict the monocular depth of the source, and then adjust the scale and offset of the depth map. After that, to mitigate the potential errors caused by deep models, we jointly optimize camera relative pose and depth map scale and offset to make the depth map and image contents well-align near the hole region. Finally, we render the reprojected source and refine it using post-processing.     
}
\vspace{-4mm}
\label{fig:pipeline_figure}
\end{figure*}

However, reference-based image inpainting is very challenging and not well explored. That is because reference-based inpainting is a photography-oriented task that has only one single reference frame as ``source", and has usually large baseline or challenging camera movements between the target and source. Therefore, the target and source images cannot be easily aligned to fill in the hole due to  issues such as parallax. To address the alignment problems, the previous state-of-the-art method TransFill~\cite{zhou2021transfill} has a strong assumption that the scenes can be blended by multiple planar structures. It clusters the predicted depth on matched feature points and utilizes multiple homographies to fill the hole. However, 
real scenes rarely consist of only a few planar surfaces, and even when they do, it can be hard to identify the relevant planar surfaces. This indicates that to better solve the problem of reference-based inpainting, \emph{it is crucial to understand the camera positions and geometry of the 3D scene, especially near the hole region} in order to find the appropriate content for hole filling. 



In this work, we propose a more principled approach that fills the hole region by explicitly estimating the 3D scene structure from two limited views. Specifically, we first estimate sparse feature correspondences, from which we derive an initial relative pose between the two views. We then predict a monocular dense depth map of the source image and determine a scale and offset that aligns the depth map with the target using a sparse 3D triangulated point cloud. To mitigate the prediction errors and improve alignment accuracy, we next jointly optimize the depth scale and offset and the relative pose using a fast differentiable 3D reprojection of the source image to the target. We synthesize a warped source image by rendering a textured source mesh with the optimized depth and target pose, and fill dis-occluded regions with single image inpainting. Lastly, we adjust the exposure, white balance, lighting, and correct any residual misalignments before pasting the result back into the hole region.


In summary, our GeoFill is the first to apply a more principled approach for reference-based inpainting, \textit{i.e.}, the first to leverage an explicit non-planar 3D scene representation given only limited two-view RGB images (no camera pose information). Compared with the previous state-of-the-art, GeoFill better handles complex 3D scene structures within the hole, wide-baseline image pairs and larger holes. Extensive experiments demonstrate that our methods achieve the best perceptual quality on various scenes in benchmarks and real user cases. 


\section{Related Work}
{\bf Image inpainting.} Traditional image inpainting models rely on hand-crafted heuristics.
Diffusion-based approaches~\cite{bertalmio2000image} propagate pixel colors from the background to the hole region. These approaches generate artifacts when the hole size is large or texture variations are significant. Alternatively, patch-based approaches~\cite{wexler2004space,barnes2009patchmatch} search for similar patches outside the hole region to complete the missing regions. Although these approaches offer high quality texture by copying texture patches, the filled regions may be inconsistent with regions around the hole due to lack of high-level structural understanding of the entire image. 

Recent deep models fill the hole by learning from large amounts of training data.
Context encoders~\cite{pathak2016context} generate semantically plausible content in the hole by encoding the surroundings. Iizuka~et~al.~\cite{iizuka2017globally} adopt two discriminators to ensure the inpainted content is both locally and globally consistent. Artifacts can be reduced along the hole boundary by filtering using partial~\cite{liu2018image} or gated~\cite{yu2019free}  convolutions. Some recent inpainting models improve the generated image quality with additional information, such as edges~\cite{nazeri2019edgeconnect}, segmentation masks~\cite{song2018spg}, and low frequency structures~\cite{ren2019structureflow,liao2020guidance}. Moreover, several papers show deep neural networks can fill holes on high resolution images~\cite{yang2017high,yi2020contextual,zeng2020high}. Despite the significant advancements in single inpainting models, filling with one single image remains fundamentally an ill-posed problem~\cite{zheng2019pluralistic}. Image inpainting with additional information is also explored in the literature, such as inpainting with stereo images~\cite{wang2008stereoscopic,bhavsar2010inpainting,baek2016multiview,ma2020learning,ma2021fov} and utilizing more than one image~\cite{thonat2016multi}. 
TransFill~\cite{zhou2021transfill} is closely related to our work: it performs reference-guided inpainting by warping a reference image with multiple homographies. 
However, due to the planar nature of homographies, TransFill has a limited ability to handle image pairs with complex 3D structures, wide baselines, or significant disocclusions.

{\bf Video Inpainting.} Classical works mainly focus on globally optimizing patch-based energies~\cite{patwardhan2005video, wexler2007space, granados2012background}. Recent work often adopts deep generative models for better inpainting performance. 
Wang \textit{et al.} introduce a data-driven framework that jointly learns temporal structure and spatial details~\cite{wang2019video}.
Onion-Peel Network (OPN) proposes to fill in the missing region progressively with the spatio-temporal attention~\cite{oh2019onion}.
Spatial-Temporal Transformer Network (STTN) adopts a deep generative model with adversarial training along the spatial-temporal dimension to mitigate blurriness and temporal artifacts~\cite{yan2020sttn}.
Note that video inpainting approaches heavily exploit the dense temporal information in the video while we only have one single reference image which is a much harder scenario. 

{\bf Two-view Geometry.} SfM establishes correspondences between two monocular frames and subsequently estimates 3D structure~\cite{hartley2003multiple,schonberger2016structure,zhou2017unsupervised,wang2021deep,ma2021transfusion}.
In classic geometric vision, it is well understood that the camera poses as well as depth for corresponding points can be computed from feature matching points alone~\cite{longuet1981computer, hartley1997triangulation}.
Traditional methods utilize hand-crafted descriptors~\cite{lowe1999object,bay2006surf,rublee2011orb} to build sparse correspondence for the subsequent fundamental matrix estimation with the 8-point algorithm~\cite{hartley1997defense}. 
Learned local features have shown great success in recent works~\cite{yi2016lift,detone2017toward,detone2018superpoint} together with the learning-based feature matching models, such as SuperGlue~\cite{sarlin2020superglue} or differentiable formations of RANSAC~\cite{brachmann2017dsac,ranftl2018deep,brachmann2019neural}. Another alternative is to directly estimate relative pose using an end-to-end pose estimation network~\cite{kendall2015posenet}. We leverage these recent advances (specifically OANet \cite{zhang2019learning}). Our method uses components inspired by SfM, however, our differentiable joint optimization stage is novel and has been carefully formulated for our task. 

{\bf Monocular Depth Estimation.} Predicting depth from a single image is an ill-posed problem. However, learning based approaches have shown impressive performance by either treating monocular depth estimation as a regression or classification task~\cite{eigen2014depth,laina2016deeper,xu2017multi,hao2018detail,xu2018structured,fu2018deep,hu2019revisiting,alhashim2018high,lee2019big,zhao2020domain, huynh2020guiding, zhao2021camera}.
Recent advances include BTS~\cite{lee2019big}, which introduces local planar guidance layers to guide the features to full resolution instead of standard upsampling layers during the decoding phase. 
DAV~\cite{huynh2020guiding} proposes to exploit the co-planarity of objects in the scene via depth-attention volume.
DPT~\cite{ranftl2021vision} leverages the high quality intermediate representation from transformers and become state-of-the-art.

\section{Method}
Suppose we are given a target image $\I_t$ with a hole $\M$ to be filled, and a reference (source) image $\I_s$ of the same scene. Our goal is to find a 3D-aware warped source image $\I_{s \rightarrow t}$ that geometrically aligns the source image to the target image which can be used to fill the hole. The final composite image can be represented as $\I_t^{\mathrm{comp}} = \I_t \odot \M + (\M_{\mathrm{single}} \I_{s \rightarrow t} + (1 - \M_{\mathrm{single}}) \odot \I_{\mathrm{single}})\odot (1 - \M)$, where $\M_{\mathrm{single}}$ is the blending map to merge the warped source with the single image inpainting result $\I_{\mathrm{single}}$. Note that ideally there should be enough contents that GeoFill can copy from the source image to the target hole region, 
\textit{i.e.} the source image $I_s$ is useful. Cases with very few target-source content pixels overlapping inside the hole will cause  GeoFill to fall back to the single image inpainting  $\I_{\mathrm{single}}$.

To compute the final warping matrix and use it to reproject the source image, as shown in Figure \ref{fig:pipeline_figure}, we propose a pipeline consisting of three stages named initialization, joint optimization, and rendering and post-processing. In the first stage, we establish sparse correspondences between $\I_s$ and $\I_t$ and estimate the relative pose $\T_{\mathrm{rel}}$ between two views. Meanwhile, we obtain a dense depth map of the source image using a pretrained deep model. Then we align the scale and offset of the predicted depth map with the sparse 3D triangulated feature points. In the second stage, we mitigate the potential errors of the initial guess of pose and depth by optimizing the related parameters so contents well-align near the hole. Finally, we render the warped image using the optimized parameters, and post-process it to address any residual spatial and color misalignments like TransFill. We will introduce each stage in the following sections.





\subsection{Initialization Stage}
\textbf{Initialize Relative Pose}
Our approach first estimates the relative pose $\T_{\mathrm{rel}}$ based on predicted sparse correspondences. We extract sparse corresponding feature points $\P_t$ and $\P_s$ between the target and source images, and compute the fundamental matrix $\F$ between $\I_s$ and $\I_t$ via the normalized 8-point algorithm~\cite{hartley1997defense} using RANSAC~\cite{fischler1981random}. From $\F$, we derive the relative pose $\T_{\mathrm{rel}}$ using the classic multi-view geometry algorithm mentioned in~\cite{hartley1997defense}.

\textbf{Initialize Dense Depth Map}
We then predict the inverse depth map on the source image using a pretrained monocular depth estimator as the cues to estimate the real source depth. Mathematically, we only need the source depth and $\T_{\mathrm{rel}}$ to compute the 3D-aware warp $\I_{s \rightarrow t}$. However, the source depth is predicted up to an unknown scale and offset by a pretrained deep model, therefore, we need to solve for these such that the initial source depth $\D_s^i$ best matches the estimated relative pose. 
Note the estimated translation in $\T_{\mathrm{rel}}$ is normalized and will not match the arbitrary scale in the predicted depth maps from pretrained deep models. As suggested in~\cite{zhao2020towards}, aligning dense depth to sparse triangulated points is much simpler than rescaling the relative pose. Therefore, we first triangulate points with the relative pose, then align the scale of depth predictions with triangulation to subsequently match the scale of the relative pose.

Specifically, a 3D triangulated point $\x$ with point $q_s \in \P_s$ and $q_t \in \P_t$ is computed as,
\begin{equation}
    \x^* = \argmin_{\x} [E({\bf r}_s, \x)]^2 + [E({\bf r}_t, \x)]^2,
\end{equation}
where ${\bf r}_s$ represents the ray shooting from the source camera center through the point $q_s$ on the image plane, ${\bf r}_t$ is the ray from the target camera following a similar analogy, and $E$ measures the Euclidean distance between two inputs. 
In this way, we compute a set of 3D triangulated points $\X$ using all matching sparse correspondences. 
In order to form the linear problem to compute the scale and offset, we first compute a sparse triangulated depth map $\D_{\mathrm{tri}}$ by projecting 3D triangulated points to source camera coordinates. 
Note $\D_{\mathrm{tri}}$ is in the same scale as the relative pose. Therefore, we correct $\D_s$ to match $\D_{\mathrm{tri}}$, which subsequently matches the scale of the relative pose. We correct $\D_s$ by estimating two scalars, the scale $s^i$ and offset $b^i$ associated with the depth map, by solving a linear least square problem. The initial depth maps are then expressed as: $\D_s^i = s^i\D_s + b^i$.

\subsection{Joint Optimization Stage}
To mitigate the effects of potential errors in sparse correspondence and depth estimation since deep models can be not robust or generalized enough, we further introduce an optimization module to improve the quality of $\I_{s \rightarrow t}$. 
We optimize the depth scale, offset, and the relative pose that jointly define $\I_{s \rightarrow t}$ in the 3D scene. 
Specifically, we convert the rotation matrix into quaternions, which leads to a total of 9 parameters to optimize. 
Both relative pose and the initial depth computed in the previous section are used as the initial guess for the optimization. 
Our optimization contains 3 different loss functions: a multiscale photometric loss $\mathcal{L}_{\mathrm{photo}}$, a feature correspondence loss $\mathcal{L}_{\mathrm{feat}}$, and a negative depth penalty $\mathcal{L}_{\mathrm{negD}}$.

{\bf Multiscale Photometric Loss $\mathcal{L}_{\mathrm{photo}}$} measures the pixel-level color difference between $\I_{s \rightarrow t}$ and the $\I_t$ outside the hole region. 
We downsample both $\I_{s \rightarrow t}$ and $\I_t$ and sum the normalized color difference across different resolutions.
Specifically, we build Gaussian pyramids on both $\I_{s \rightarrow t}$ and $\I_t$ using an RGB representation for the source image and an alpha-premultiplied RGBA representation on the target image to incorporate the hole region properly into the target image.
Computing a multi-scale photometric loss within each iteration is obviously computationally expensive. Moreover, the optimization might also get trapped into the local minima associated with the finest resolution due to poor initialization~\cite{zhu2015face}. 
To accelerate the computation speed and find better solutions, we adopt a coarse-to-fine optimization strategy, which means we first compute photometric loss on the most coarse level and move to the finer level once the convergence criteria at the current pyramid level are met.
Additionally, instead of building a 3D triangle mesh and rendering from the target view at each iteration, we use a much more efficient differentiable 3D reprojection to find a warping field that computes $\I_{s \rightarrow t}$ from $\I_s$ with bilinear interpolation. Mathematically, we have:
\begin{equation}
   \I_{s \rightarrow t} = \mathrm{bilinear}(\I_s, \mathrm{reproj}(\K, \T_{\mathrm{rel}}, \D_s^o)),
\end{equation}
where $\mathrm{reproj}()$ represents the reprojection operation. The photometric loss at a given resolution with the pyramid is:
\begin{equation}
    \mathcal{L}_{\mathrm{photo}} = \frac{1}{|\M|} \sum \W \odot ||\I_{s \rightarrow t} \odot \M - I_t \odot \M||^2.
\end{equation}
Here $\W$ is a pixel importance weight map discussed shortly.

{\bf Feature Correspondence Loss $\mathcal{L}_{\mathrm{feat}}$} computes the distance between reprojected matching feature points in the source images and the target images.
We use the $\mathrm{reproj}()$ operator on all 2D image coordinates in $\P_s$ to get another set $\P_{s \rightarrow t}$. Then, we compute the average distance between $\P_{s \rightarrow t}$ and $\P_t$. 
However, the average distance of all points is very sensitive to outliers, i.e., very few outliers dominate the loss function. 
To reduce the effects of the outliers on the loss function, we adopt the general robust loss function from~\cite{barron2019general}. The general form of the loss function is:
\begin{equation}
    f(x, \alpha, c) = \frac{|\alpha - 2|}{\alpha} \left( \left(\frac{(x/c)^2}{|\alpha-2|}+1 \right) ^{\alpha/2}-1 \right),
\end{equation}
where $\alpha$ and $c$ are the shape and scale parameters, respectively. In our experiments, we set $\alpha=-2$ and $c=10$.
Then, feature correspondence loss is written as:
\begin{equation}
    \mathcal{L}_{\mathrm{feat}} = \frac{1}{|P_{s \rightarrow t}|} \sum\limits_{m=0}^{|P_{s \rightarrow t}|} \W(\q^m_{t}) f(||\q^m_{s \rightarrow t} - \q^m_{t}||, \alpha, c),
\end{equation}
where $f$ is the general robust loss function, $\q^m_{s \rightarrow t}$ and $\q^m_{t}$ is the $m^{th}$ point in $P_{s \rightarrow t}$ and $P_t$, respectively.

{\bf Negative Depth Penalty $\mathcal{L}_{\mathrm{negD}}$} aims to penalize negative values in the remapped depth. Although depth predictions have arbitrary scale and offset, they should have all positive values, meaning that a fragment of geometry associated with a pixel should never move behind the camera. Mathematically, we adopt a hinge loss function:
\begin{equation}
    \mathcal{L}_{\mathrm{negD}}= \sum \max\{0, -\D_s^o\}.
\end{equation}
Our final objective function is written as: $l = \lambda_1 \mathcal{L}_{\mathrm{photo}} + \lambda_2 \mathcal{L}_{\mathrm{feat}} + \lambda_3 \mathcal{L}_{\mathrm{negD}}$, where $\{ \lambda_j \}$ are weights. The convergence criteria is discussed in the supplemental material.

{\bf Pixel Importance Weight Map \W.} When computing the photometric loss or feature correspondence loss, we assign weights to each pixel to replace uniform weighting. It encourages our optimization to better align the warped source with the target image both locally and globally. 

The first type of weighting strategy is hole-distance weighting $\W_h$. We encourage the optimization to focus on the regions close to the hole boundary since those pixels are more important for filling the hole regions. To achieve this, we apply the distance transform to the hole image and obtain a distance map $\M_{h}$, where each pixel records the Euclidean distance to the closest boundary pixel of the hole $\M_{h}$. We compute the weight at each pixel using a Gaussian function $\W_{h} = \exp(- \M_{h}^2 / 2\sigma^2)$, where $\sigma$ is a hyperparameter that adjusts how weights change w.r.t distance to the hole boundary. 

The second type is the edge-based weighting $\W_e$. This is because high-gradient edge regions are more salient while checking the alignment quality so we intend to give strong edges larger weights. We compute a multi-scale Canny edge map by first applying Gaussian blur on $\I_t$ with N different kernel sizes, running the Canny edge detector~\cite{canny1986computational}, and dilating each edge to get $\{e_1, e_2, \cdots, e_N\}$. Our pixel-level edge-based weight map becomes $\W_{e} = \sum_{k=1}^N e_{k} / \sum_p e_{k}(p)$, where the inner sum is over spatial coordinates. Our overall weighting map is $\W = \W_{h} \odot \W_{e}$.

\subsection{Rendering and Postprocessing Stage}
{\bf Mesh Rendering.} After optimization, we find $\T_{\mathrm{rel}}^{o}$, $s^{o}$, and $b^{o}$, the camera relative pose and depth map scale and offset, respectively. However, computing $\I_{s \rightarrow t}$ using depth reprojection always has some gaps between valid pixels due to it relying on a forward warping for efficiency in the inner loop of the optimization.  One way to overcome this problem is interpolation but that has two disadvantages. First, the reprojected image is sparse and for regions with multiple layers of depth, pixels for a far-away depth layer might splat between pixels of a closer depth layer, which could result in interpolations that do not fully remove hidden surfaces. Additionally, interpolation cannot distinguish between holes due to disoccluded regions or simple gaps between the pixels due to the forward warping.

To address the above problems, we choose to render a textured mesh to get the final $\I_{s \rightarrow t}$. We first build a triangle mesh with a regular grid from the source view. The mesh vertices are computed by projecting the optimized depth $\D_s^{o}$ to the 3D space and the texture is the RGB colors of the source image. After obtaining the mesh, we drop the edges around depth discontinuities. We adopt a simplified version of the footprint algorithm~\cite{westover1990footprint} by comparing the depth values between connected vertices. We drop the edge between two vertices $v_i$ and $v_j$ if $
    \frac{2|d(v_i) - d(v_j)|}{d(v_i) + d(v_j)} > \epsilon_{\mathrm{edge}}$, where $d(v_i)$ is the depth value of the vertex $v_i$ and $\epsilon_{\mathrm{edge}}$ is a predefined threshold. After building the triangle mesh, we render the target view with $\T_{\mathrm{rel}}^o$. Note that we also normalize the mesh to the unit size before rendering. The textured mesh densely fills in pixels and removes hidden surfaces. It also allows us to use the rendered alpha channel to find pixels where there is no ray intersection with the mesh, which represent disoccluded regions near depth discontinuities or regions outside the photo: these are later filled by single image inpainting. We use PyTorch3D~\cite{ravi2020accelerating} as our renderer.

{\bf Refinement and Merging.} With the rendered image, we apply the color-spatial transformation (CST) module from TransFill to further improve any small residual spatial misalignments and correct color and exposure differences.
Lastly, we merge the output from CST with results from the single image inpainting model as in TransFill to handle the disocclusions and regions outside the photo.

\begin{figure*}[t]
\centering
\includegraphics[width=0.99\textwidth]{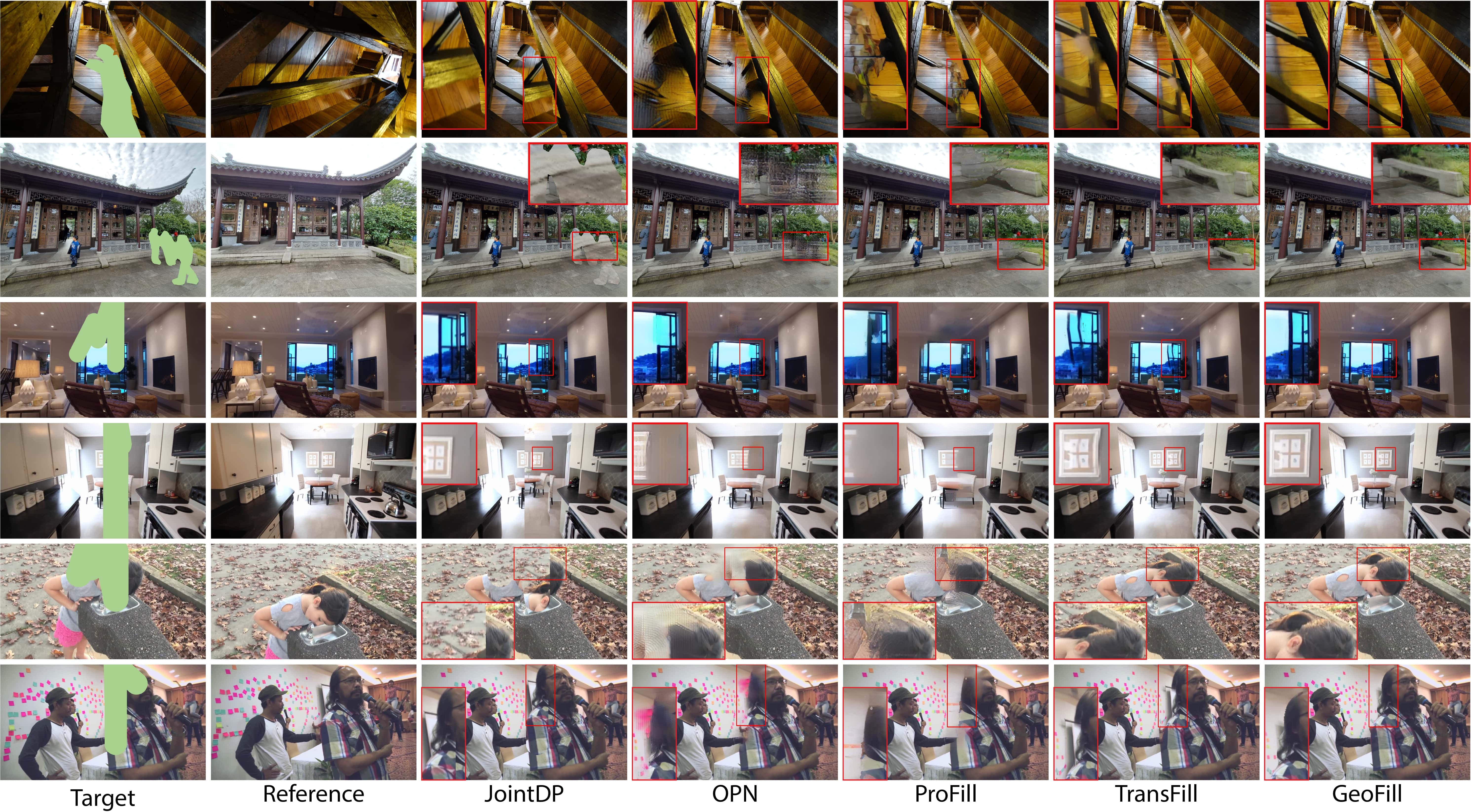}
\vspace{-2ex}
\caption{Qualitative comparison of GeoFill against other baselines on user-provided images (top 2 rows), RealEstate10K (mid 2 rows), and MannequinChallenge dataset (last two rows).
}
\vspace{-4mm}
\label{fig: visualizations}
\end{figure*}

\section{Experiments}
{\bf Datasets.} 
No large-scale image-based dataset for reference-based inpainting is available. So we follow TransFill to randomly sample multiple image pairs from video-based datasets because it is easier to simulate user behaviours and analyze the target-source difference (or camera view changes). During evaluation we only use one single reference frame. This is different from what video inpainting works did. We follow DeepFillv2~\cite{yu2019free} to generate random free-form brush stroke masks 
 and evaluate GeoFill and other baselines on the following datasets.\\
\noindent\textit{RealEstate10K} \cite{zhou2018stereo}: It contains a diverse collection of YouTube video sequences shot from a moving camera for both indoor and outdoor scenes. Each video clip contains a variety of views of the same scene. We randomly sample 500 videos and select one pair of images in each video sequence with a specific frame difference (FD). Specifically, we sample FD=25, 50, and 75 to build three different sets with a resolution of $720 \times 1280$ while automatically filtering out image pairs without enough overlapping content inside the hole by checking the number of matching sparse features. Note that our filtering mechanism is only for the purpose of simulating user behaviors and removing non-useful image pairs. In practice, we believe users could simply check overlap between photos by visual inspection. \\
\noindent\textit{MannequinChallenge}~\cite{li2019learning}: This is a challenging dataset with video sequences shot from a hand-held camera freely roaming in a scene of people with frozen poses. This dataset contains more than 170K frames and corresponding camera poses derived from about 2k YouTube videos. The camera motion in this dataset is more extreme and scene complexity is much higher due to diverse frozen human poses and rich background objects. We reduce the sampling FD to ensure enough overlap between image pairs. Similar to the previous dataset, we randomly sample 3 subsets with FD=10, 20, 30, where each contains 500 image pairs of $720 \times 1280$. \\
\noindent\textit{Real User-provided Images}: We also evaluate our approach on the real user-provided images like TransFill to validate the generalization ability and practicability.

{\bf Baselines.} In addition to TransFill, which is directly related to our work,
we compare our approach against several different types of baselines to evaluate final inpainting performance.
The first type is state-of-the-art video completion models. OPN~\cite{oh2019onion} achieves high quality inpainting results with spatio-temporal attention. STTN~\cite{yan2020sttn} proposes to optimize a spatial-temporal adversarial loss function. 
In addition, we compare against state-of-the-art single-image inpainting methods including  ProFill~\cite{zeng2020high} and CoModGAN~\cite{zhao2021large}. We use their off-the-shelf pretrained weights directly since they are both trained on more diverse scenes in Places2 \cite{zhou2017places} than RealEstate10K and have strong generalization ability.
Lastly, we compare a two-view SfM based approach~\cite{zhao2020towards} --- which we refer to as JointDP --- by warping the source image with the jointly estimated relative pose and depth using dense correspondence. 
To ensure fairness in the comparison, we use the same depth predictor and rendering process as in GeoFill while keeping all other settings the same as in the original work.

{\bf Implementation Details.} We follow TransFill to extract SIFT features~\cite{lowe2004distinctive} and feed them into OANet~\cite{zhang2019learning} to reject outliers and establish correspondences. This combination was already proven to be quite robust.  A
better matching strategy could always be adopted to further improve results while leaving the rest of our framework intact. Our pretrained monocular depth predictor is DPT~\cite{ranftl2021vision}. 
No ground-truth camera intrinsic information is required to run our approach. 
We use fixed camera intrinsic parameters for all images by setting focal length to 750 and principal point to the image center. An ablation study with different focal lengths is included in the supplemental material. We use the pretrained CST module from TransFill~\cite{zhou2021transfill} without finetuning. This module generalizes well to MannequinChallenge and user provided images. Our pipeline is implemented with PyTorch~\cite{paszke2019pytorch} and we choose DiffGrad~\cite{dubey2019diffgrad} as our optimizer due to its fast convergence speed.
In the optimization step, we use a constant learning rate $10^{-2}$ and the maximum number of iteration is set to $10^4$. 
The loss weights $\lambda_1$, $\lambda_2$, and $\lambda_3$ are 10, 10, 0.5, respectively. 
In the coarse-to-fine optimization strategy, the number of pyramid levels is 4 and the maximum number of cumulative iterations at each level from coarse to fine are $4 \times 10^3$, $7 \times 10^3$, $9 \times 10^3$, $10^4$. We set the $\sigma$ in hole-based weighting to 192 pixels. In edge-based weight, we compute 4 different Canny edge maps and dilate each of them with a kernel size equal to 4. 
In mesh rendering, the edge threshold $\epsilon_{\mathrm{edge}}$ is $4 \times 10^{-2}$.

\begin{table*}[t]
\setlength{\abovecaptionskip}{0pt}
\centering
\footnotesize
\caption{Quantitative comparisons of GeoFill against other baselines on both RealEstate10K and MannequinChallenge.}
\resizebox{\textwidth}{!}{
\begin{tabular}{|r||c|c|c||c|c|c|}
\hline
&\multicolumn{3}{c||}{RealEstate10K: PSNR $\uparrow$/ SSIM $\uparrow$  / LPIPS $\downarrow$ }&\multicolumn{3}{c|}{MannequinChallenge: PSNR $\uparrow$/ SSIM $\uparrow$  / LPIPS $\downarrow$}\\\hline
Model&FD=25&FD=50&FD=75 &FD=10&FD=20&FD=30 \\ \hline\hline
JointDP \cite{zhao2020towards} & 22.46 / 0.9469 / 0.1011 & 21.76 / 0.9457 / 0.1063 & 20.89 / 0.9423 / 0.1122 & 20.13 / 0.9346 / 0.1087 & 19.52 / 0.9290 / 0.1195 & 19.38 / 0.9315 / 0.1177 \\\hline
OPN \cite{oh2019onion} & 28.41 / 0.9684 / 0.0525 & 27.80 / 0.9669 / 0.0570 & 26.91 / 0.9634 / 0.0624 & 25.63 / 0.9628 / 0.0605 & 24.92 / 0.9584 / 0.0698 & 24.84 / 0.9591 / 0.0702 \\\hline
STTN~\cite{yan2020sttn} & 28.83 / 0.9696 / 0.0710 & 28.26 / 0.9697 / 0.0721 & 27.59 / 0.9680 / 0.0751 & 25.60 / 0.9623 / 0.0803 & 25.09 / 0.9602 / 0.0865 & 24.94 / 0.9613 / 0.0844 \\\hline
ProFill~\cite{zeng2020high} & 27.45 / 0.9642 / 0.0775 & 27.67 / 0.9654 / 0.0755 & 27.37 / 0.9639 / 0.0768 & 25.04 / 0.9589 / 0.0808 & 25.02 / 0.9582 / 0.0836 & 25.22 / 0.9599 / 0.0810 \\\hline
CoModGAN~\cite{zhao2021large} & 26.02 / 0.9594 / 0.0703 & 26.14 / 0.9607 / 0.0686 & 25.88 / 0.9596 / 0.0697 & 23.39 / 0.9504 / 0.0770 & 23.14 / 0.9486 / 0.0808 & 23.36 / 0.9503 / 0.0791 \\\hline
TransFill~\cite{zhou2021transfill} & 32.03 / 0.9764 / \bf {0.0461}
 & 30.64 / 0.9732 / 0.0540 & 29.24 / 0.9694 / 0.0608 & 28.01 / 0.9680 / 0.0569 & 26.56 / 0.9628 / 0.0688 & 26.17 / 0.9632 / 0.0701
 \\\hline
GeoFill (Ours) & \textbf{32.57 / 0.9775 /} 0.0467 & \bf{31.47 / 0.9748 / 0.0525} & \bf{30.43 / 0.9717 / 0.0581} & {
\bf 28.85 / 0.9702 / 0.0553} & {\bf 27.72 / 0.9658 / 0.0652} & {\bf 27.44 / 0.9664 / 0.0665} \\
\hline
\end{tabular}}
\vspace{-6mm}
\label{exp:quant}
\end{table*}

\subsection{Quantitative Results}
The quantitative results comparing our approach with other baselines are shown in Table~\ref{exp:quant}. We report the PSNR, SSIM and LPIPS\cite{zhang2018perceptual} on the RealEstate10K and the MannequinChallenge datasets. 
Single image inpainting models are not competitive enough for image pairs with larger scale differences and wider baselines. 
Video inpainting approaches also show bad performance due to the lack of dense temporal information and multiple reference frames.
JointDP is based on optical flow and is not able to accurately estimate parameters like the camera pose needed to correctly align the image pairs. Our method demonstrates superiority over TransFill because we have a better understanding of the 3D structures of the scenes, and better leverage the depth estimation. 
Note that GeoFill has higher performance gain over TransFill on MannequinChallenge than on RealEstate10K dataset. TransFill relies on the fusion module to handle multiple homographies for the final hole filling and it is less robust on the images different from the training data. 
In contrast, GeoFill only has a single proposal to merge during inpainting, which also greatly reduces blending artifacts that arise from multiple proposals. Therefore, our GeoFill is robust and stable on image pairs with even larger frame differences.

\subsection{Qualitative Results}
Figure~\ref{fig: visualizations} shows the visual comparisons with other baseline algorithms on the user-provided images, the RealEstate10K and the MannequinChallenge dataset. JointDP utilizes estimated optical flow for the initial matching, thus the results fail to obtain accurate depth and camera pose if the baseline of the image pair is wide. The contents inside the hole are often misaligned with the target image. 
The original OPN uses five reference frames to achieve a more efficient non-local matching among frames, but a single reference frame makes the results less visually-pleasing. 
ProFill is not able to take advantage of the reference image contents, and TransFill usually has blending artifacts or content misalignment issues when objects inside the hole regions occupy multiple depth planes. However, GeoFill avoids the blending artifacts by using one single proposal, and aligns the objects well by understanding the camera poses and reconstructing the 3D scene from two images.

\subsection{Ablation Study}
In the following sections, we study how each optimization loss function and the pixel importance weight maps contribute to the final results. Additionally, we present experimental results of TransFill and GeoFill on cases with larger holes and their alignment accuracy without CST.
All experiment results in the following section are reported for the RealEstate10K FD=50 subset unless specified.
We also include more in-depth analyses of the joint depth pose optimization module on RealEstate10K and ScanNet~\cite{dai2017scannet}, the timing of each module in our system pipeline, the convergence criteria of the optimization step, 
and more visualizations in the supplemental material.

\begin{table}[t]\setlength{\tabcolsep}{13pt}
\setlength{\abovecaptionskip}{0pt}
\centering
\footnotesize
\caption{Ablations on the objective functions in the joint optimization stage of GeoFill.}
\resizebox{\columnwidth}{!}{
\begin{tabular}{|c|c|c|c|c|c|}
\hline
$\mathcal{L}_{\mathrm{photo}}$&$\mathcal{L}_{\mathrm{feat}}$&$\mathcal{L}_{\mathrm{negD}}$&PSNR$\uparrow$&SSIM$\uparrow$&LPIPS$\downarrow$\\ \hline\hline
\checkmark&\checkmark&\checkmark & \bf{31.47} & \bf{0.9748} & \bf{0.0525} \\\hline
\xmark&\checkmark&\checkmark & 31.19 & 0.9742 & 0.0533 \\\hline
\checkmark&\xmark&\checkmark & 30.88 & 0.9734 & 0.0554 \\\hline
\checkmark&\checkmark&\xmark & 31.23 & 0.9742 & 0.0532 \\
\hline
\end{tabular}
}
\label{exp:abla_optim_components}
\vspace{-2ex}
\end{table}

\begin{table}[t]\setlength{\tabcolsep}{13pt}
\setlength{\abovecaptionskip}{0pt}
\centering
\footnotesize
\caption{Ablations on the pixel importance weight map $\W$.}
\resizebox{\columnwidth}{!}{
\begin{tabular}{|c|c|c|c|c|}
\hline
Hole $\W_h$&Edge $\W_e$&PSNR$\uparrow$&SSIM$\uparrow$&LPIPS$\downarrow$\\ \hline\hline
\checkmark&\checkmark& \bf{31.47} & \bf{0.9748} & \bf{0.0525} \\\hline
\checkmark&\xmark & 31.20 & 0.9740 & 0.0534 \\\hline
\xmark&\checkmark & 31.12 & 0.9739 & 0.0539 \\\hline
\xmark&\xmark & 30.95 & 0.9734 & 0.0552 \\
\hline
\end{tabular}
}
\label{exp:abla_importance_weighting}
\vspace{-4ex}
\end{table}

{\bf Optimization: Objective Functions.} 
We claim that all the objective functions used in the optimization process contribute to a higher perceptual and reconstruction quality as shown in Table~\ref{exp:abla_optim_components}. 
Comparing GeoFill without photometric loss against without feature correspondence loss, the feature loss contributes to the performance the most. 
We find that photometric loss by itself may get distracted by local textures and fall into local minima, such that it achieves lower RGB errors on average but ignores the global structure. However, using the photometric loss does help improve alignments.
Lastly, GeoFill without negative depth penalty still has a performance drop, suggesting negative depth penalty is able to prevent corner cases where depth scale or offset estimates are non-robust.

{\bf Optimization: Pixel Importance Weight Map.} We present results of GeoFill with different combinations of pixel importance weight maps in the optimization in Table~\ref{exp:abla_importance_weighting}. This suggests that hole-distance weighting $\W_h$ which puts higher weights around the hole lead to better local alignment around the hole. Edge-based weighting $\W_e$ also helped by matching strong edges within the image. 
Using a uniform weighting map leads to the worst performance, indicating the effectiveness of the weighting maps. 

\begin{figure}[t]
\centering
\includegraphics[width=0.99\linewidth]{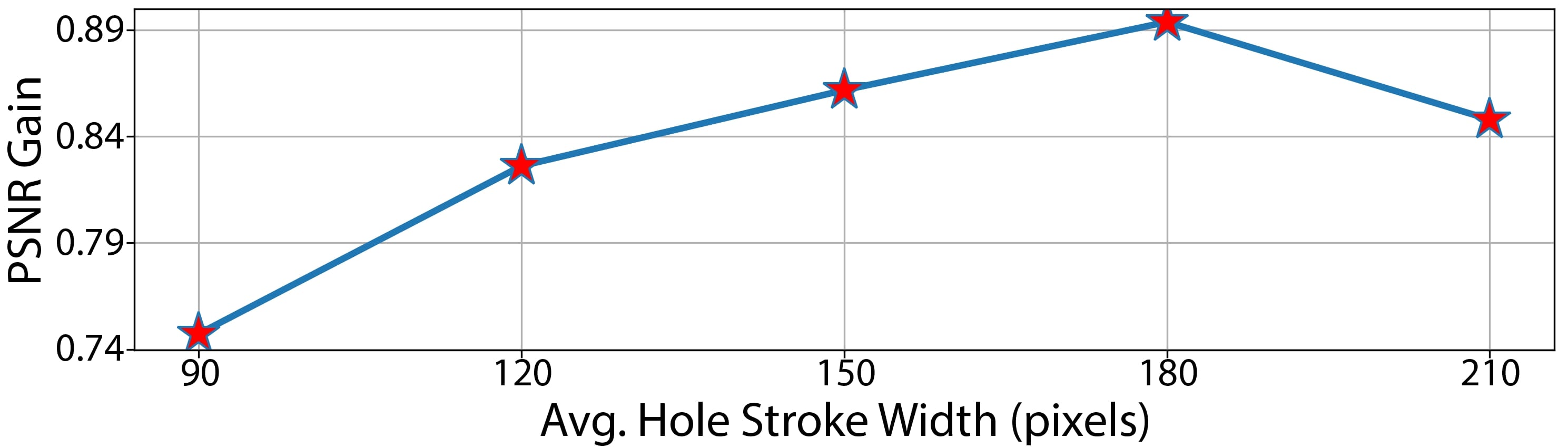}
\caption{\small Performance gain of our method compared to TransFill w.r.t the average hole size. GeoFill has a greater advantage when the hole is larger.}
\label{fig: hole_size_abl}
\end{figure}

{\bf Performance w.r.t Hole Size.} This ablation study aims to examine the performance of our approach against TransFill under more difficult settings. 
As the hole becomes larger, inevitably we have fewer matching points, which makes $\I_{s \rightarrow t}$ hard to align with $\I_t$. We generate holes with different average stroke widths ranging from 90 to 210 pixels. 
As shown in Figure~\ref{fig: hole_size_abl}, GeoFill has an increasing performance gain over TransFill until the hole average stroke width reaches 180 pixels. As the hole size grows, the complexity of the content in the hole also increases. In other words, we are more likely to encounter a growing number of depth layers, a more complicated objects layout, and a higher chance of occlusions due to camera translation in a larger hole. These problems are harder for homography-based models, therefore, GeoFill has a greater advantage when the hole is larger. The drop in performance gain at the end of the curve can be because there are not enough matching points for GeoFill to infer an accurate 3D structure.

{\bf Initial Alignment Comparisons against TransFill.} We adopted the CST module from TransFill to adjust auto exposure, lighting conditions, and potential remaining misalignments. In this experiment, we analyze the performance of using the aligned images directly, e.g. without using CST for both models. We retain the multiple homographies used by TransFill, drop the CST module, and keep the merging module so TransFill can merge its different regions. Table~\ref{exp:abla_no_cst} shows that GeoFill maintains better performance, dropping 2.12 in PSNR, while TransFill drops 4.62 without CST. This experimentally validates the proposal from GeoFill is much more accurate than TransFill by better leveraging the depth.


\begin{table}[t]\setlength{\tabcolsep}{13pt}
\setlength{\abovecaptionskip}{0pt}
\centering
\footnotesize
\caption{Ablations on Initial alignment comparisons of our method compared to TransFill without the CST Module.}
\begin{tabular}{|c|c|c|c|}
\hline
Model &PSNR$\uparrow$&SSIM$\uparrow$&LPIPS$\downarrow$\\ \hline\hline
GeoFill & \bf{31.47} & \bf{0.9748} & \bf{0.0525} \\\hline
GeoFill (no CST) & 29.35 & 0.9688 & 0.0579 \\\hline
TransFill & 30.64 & 0.9732 & 0.0540 \\\hline
TransFill (no CST) & 26.03 & 0.9598 & 0.0742 \\
\hline
\end{tabular}
\label{exp:abla_no_cst}
\vspace{-4mm}
\end{table}

\begin{figure}[t]
\centering
\includegraphics[width=0.93\linewidth]{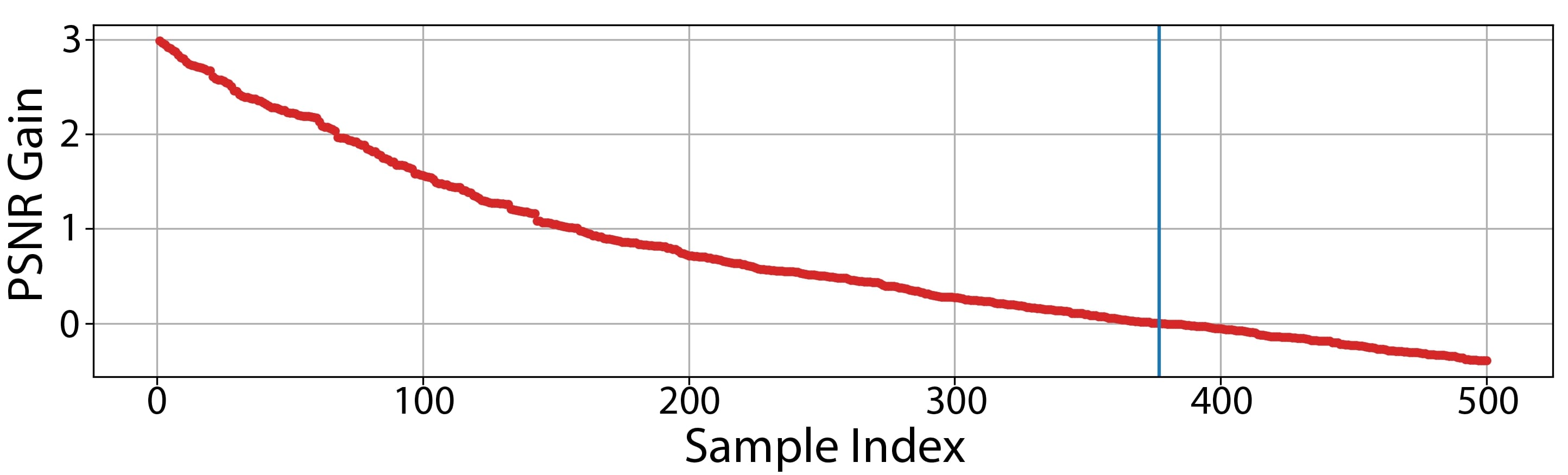}
\caption{\small Per-sample performance gain of ours compared to TransFill. The blue vertical line separates positive and negative PSNR gain.}
\label{fig: per_sample_gain}
\vspace{-2mm}
\end{figure}

{\bf Per-Sample Improvement Study.} This ablation is designed to break down the numbers shown in Table~\ref{exp:quant} by comparing the performance of GeoFill against TransFill over each individual sample. 
We sort and plot the per-image PSNR difference over TransFill, shown in Figure~\ref{fig: per_sample_gain}. Specifically, PSNR difference is computed as GeoFill PSNR $-$ TransFill PSNR, therefore a positive PSNR gain indicates GeoFill is better. 
GeoFill improves performance on the majority of samples, specifically for around 75.4\% of the entire subset. Moreover, the PSNR gains when GeoFill outperforms are much greater: up to 3 dB, as opposed to the PSNR losses when TransFill outperforms. 

\section{Discussion, Limitations, and Conclusion}
{\bf Limitations and Future Work.} GeoFill inpaints the hole with one image pair with no auxiliary pose or depth information from sensors, therefore, our pipeline may not work well when the quality of feature matching points is poor, e.g., matching points are inaccurate or too few. Under these cases, our relative pose and triangulated points can be inaccurate, which may be hard for optimization to correct. Additionally, our pipeline is also sensitive to depth prediction quality: artifacts such as blurry depth discontinuities or wrong order of depth planes can lead to potential bad inpainting results. Future work might mitigate these problems by jointly reasoning about monocular depth and the stereo cues established by triangulation. In the optimization, we used a 3D reprojection based on forward warping because it is much faster than rendering a triangle mesh even though it does not fully remove hidden surfaces in the rare cases where a mesh occludes itself: this could be addressed in future work by testing and pruning those splatted points. GeoFill utilizes the CST module to adjust auto exposure and lighting condition changes, which still suffers when the scene environment changes drastically, e.g., day to night, spring to fall. Future work could better address these by incorporating specialized lighting estimation (e.g.,~\cite{zhang2019all}) and relighting modules. One last limitation of GeoFill is that our pipeline only handles static scenes. Any dynamic objects, e.g., walking people or moving cars, not in the hole could lead to bad relative pose estimation and our optimization may cause misalignments under such cases.
A simple solution would be masking out objects that are likely to move such as people and cars, but how to correctly identify all moving objects is still an open question.

{\bf Conclusion.} In this work, we proposed a novel reference-based inpainting approach that achieves state-of-the-art performance. Our method relies on off-the-shelf depth estimation and sparse correspondence methods, and thus we can expect its performance to improve if these modules improve in performance.
Compared to previous homography-based work, GeoFill is better at image pairs with large camera motions and scenes with complicated structures in the hole. 
GeoFill ensures accurate alignment between the filling content and the hole by jointly optimizing depth predictions and relative poses without explicitly knowing camera intrinsics. 
Moreover, our approach inpaints the hole in a more principled way by utilizing a single proposal without heuristic planar assumptions.


{\small
\bibliographystyle{ieee_fullname}
\bibliography{egbib}
}

\clearpage
\appendix
\addcontentsline{toc}{section}{Appendices}

\begin{center}
{\bf \Large Appendices}
\end{center}

\emph{
In the supplementary document, we provide additional ablation studies to further support our findings, as well as details of our experiments and more visualizations. Below is the outline.
}
\begin{itemize} [noitemsep, topsep=-1pt, leftmargin=*]
\item 
    {\bf Section~\ref{ssec:convergence}: Convergence criteria.}
    We describe the convergence criteria of our optimization step in detail.
\item
    {\bf Section~\ref{ssec:init vs optim}: Joint optimization evaluation: inpainting performance.} 
    We demonstrate the importance of the joint optimization module by comparing the performance of GeoFill with parameters estimated before and after the joint optimization. 
\item
    {\bf Section~\ref{ssec:depth pose eval}: Joint optimization evaluation: depth and pose accuracy.} 
    We further evaluate the performance of the optimization module by measuring the accuracy of depth maps and camera poses individually against ground-truth using ScanNet.
\item
    {\bf Section~\ref{ssec:focal_study}: Performance w.r.t intrinsic parameters.} 
    Quantitative comparisons of our approach with different focal lengths. 
\item 
    {\bf Section~\ref{ssec:no cst}: Further analyses of initial alignments.} 
    We provide additional qualitative and quantitative evaluations of GeoFill and TransFill without the CST module. 
\item
    {\bf Section~\ref{ssec:hole ablation vis}: Visualizations of hole ablation study.} 
    We include qualitative comparisons between GeoFill and TransFill under various hole sizes. 
\item
    {\bf Section~\ref{ssec:timing}: Average running time of GeoFill.} 
    We report the average running time of each step in GeoFill.
\item 
    {\bf Section~\ref{ssec:user_study}: User study against other baselines.} 
    A user study of GeoFill against OPN, ProFill, and TransFill.
\item 
    {\bf Section~\ref{ssec:appear}: Handling appearance changes from camera movement.} 
    We show qualitative results of GeoFill handling some common appearance changes due to camera movement.  
\item 
    {\bf Section~\ref{ssec:failure}: Failure cases.} 
    Visual examples of failure cases of GeoFill. 
\item
    {\bf Additional Visual Results.} 
    We include more inpainting results in Fig.~\ref{fig: suppl_vis1} and \ref{fig: suppl_vis2}.
\end{itemize}

\section{Convergence Criteria}
\label{ssec:convergence}
The convergence criteria define when the  optimization should stop. Our optimization halts the loop at a given scale and continues to the next scale if the following condition is met or the predefined maximum number of iteration is achieved. The formula below measures the objective function value changes within the last $m$ iterations. 
\begin{equation}
    \label{optim_cond}
    \epsilon_i = \frac{|\sum_{i-(m/2)-1}^{i} l_{i} - \sum_{i-(m/2)}^{i-m-1} l_{i}|}{\sum_{i-(m/2)-1}^{i} l_{i}},
\end{equation}
where $i$ represents the $i^{th}$ iteration. If $\epsilon_i$ is smaller than a predefined $\epsilon_{opt}$, we assume the objective function has converged. Since we adopt a coarse-to-fine optimization strategy, we check the same condition at every level of the pyramid. In other words, we move to the finer scale level only if Eqn.~\ref{optim_cond} is met or maximum number of iterations at the current level is reached. We also keep track of the optimal parameters at each level and use them as the initialization in the next level. 
In practice, we set the convergence threshold $\epsilon_{opt}$ to $10^{-6}$ for all levels. The number of loss values to track in computing convergence criteria is $m=10$. 

\section{Joint Optimization Evaluation: Inpainting Performance}
\label{ssec:init vs optim}
We show the importance of our optimization module by comparing the performance of GeoFill with initial estimated parameters and optimized parameters. As shown in Table~\ref{tab:init vs optim}, GeoFill with optimized parameters has substantially better performance. Initial parameters are computed from SIFT and pretrained models such as OANet, which can make erroneous predictions, especially for image pairs with holes. Experimental results demonstrate our optimization module successfully mitigates such errors and improves the inpainting performance. 

\section{Joint Optimization Evaluation: Depth, Pose Accuracy}
\label{ssec:depth pose eval}
In this section, we further demonstrate the effectiveness of our joint optimization step by measuring the depth and pose accuracy before and after the optimization. Since both RealEstate10K and MannequinChallenge do not have ground-truth labels, we choose the ScanNet~\cite{dai2017scannet} dataset which comes with ground-truth camera poses and depth maps. 
We randomly sampled 75 pairs of images with approximately 30 frame difference. We generate random holes in the same manner as described in our main paper line 388. Each image pair comes from a unique scene in the dataset. Note that ScanNet includes images with heavy motion blur, which we manually filtered out.
We evaluate depth and relative camera pose \emph{separately} by providing the ground-truth for one of these (depth or camera pose) when evaluating the accuracy of the other one.
For example, when evaluating the accuracy of depth maps, we first follow the same pipeline described in the main paper. Then, instead of estimating the relative pose, we provide the ground-truth camera pose and evaluate the accuracy of the depth map determined by our pipeline before and after the optimization. Note that we only optimize scale and offset when evaluating depth accuracy. 
A similar analogy applies when evaluating the accuracy of camera poses: we provide the ground truth depth map to our pipeline and then evaluate the accuracy of the relative pose before and after optimization. 
For pose evaluation, we report the geodesic errors~\cite{chen2021wide} for both rotations and translation directions. 
For depth evaluation, we follow the commonly adopted metrics used in the literature~\cite{eigen2014depth,fu2018deep}. As shown in Table~\ref{tab:pose_eval} and Table~\ref{tab:depth_eval}, both depth and pose errors are significantly reduced after the optimization module, which demonstrates the ability of the optimization module to find more accurate depth and poses in our challenging case where the images have holes. 

\begin{table}[t]\setlength{\tabcolsep}{13pt}
\setlength{\abovecaptionskip}{0pt}
\centering
\footnotesize
\caption{Quantitative comparison of our method with initially estimated parameters and optimized parameters.}
\begin{tabular}{|c|c|c|c|}
\hline
Model &PSNR$\uparrow$&SSIM$\uparrow$&LPIPS$\downarrow$\\ \hline\hline
GeoFill (optim) & \bf{31.47} & \bf{0.9748} & \bf{0.0525} \\\hline
GeoFill (init) & 30.66 & 0.9719 & 0.0548 \\\hline
\end{tabular}
\label{tab:init vs optim}
\vspace{-2mm}
\end{table}

\begin{table}[t]\setlength{\tabcolsep}{6pt}
\setlength{\abovecaptionskip}{0pt}
\centering
\footnotesize
\caption{Relative camera pose evaluation of initial guess and our optimized results, where $R$ and $t$ represent the rotation and translation, respectively.}
\begin{tabular}{lccccc}
\hline
\toprule
& \multicolumn{2}{c}{$R\downarrow$} && \multicolumn{2}{c}{${t\downarrow}$} \\
\cmidrule{2-3} \cmidrule{5-6}
& mean ($^\circ$) & med ($^\circ$) && mean ($^\circ$) & med ($^\circ$) \\
\midrule 
GeoFill (optim) & {\bf 1.588} & {\bf 1.062} && {\bf 3.688}  & {\bf 3.457}  \\\hline
GeoFill (init) & 7.378 & 7.807 && 11.861 & 11.096
\\\hline
\end{tabular}
\vspace{-4mm}
\label{tab:pose_eval}
\end{table}

\begin{table}[t]\setlength{\tabcolsep}{6pt}
\setlength{\abovecaptionskip}{0pt}
\centering
\footnotesize
\caption{
Depth evaluation of our initial guess and optimized results. The evaluation metrics include absolute relative difference (Abs$^r$), squared relative difference (Sq$^r$), root mean squared log error (RMS-log), and accuracy with a relative error threshold of $\delta^k < 1.25^k$, k = 1, 2.
}
\begin{tabular}{| c | c | c | c | c | c |}
\hline
Models & Abs$^r\downarrow$ & Sq$^r\downarrow$ & RMS-log$\downarrow$ & $\delta^1\uparrow$ & $\delta^2\uparrow$ \\
\hline\hline
GeoFill (optim) & {\bf .258} & {\bf .233} & {\bf .302} & {\bf .683} & {\bf .851} \\\hline
GeoFill (init) & .372 & .766 & .403 & .609 & .788 \\\hline
\end{tabular}
\label{tab:depth_eval}
\end{table}

\begin{figure}[t]
\centering
\includegraphics[width=0.99\linewidth]{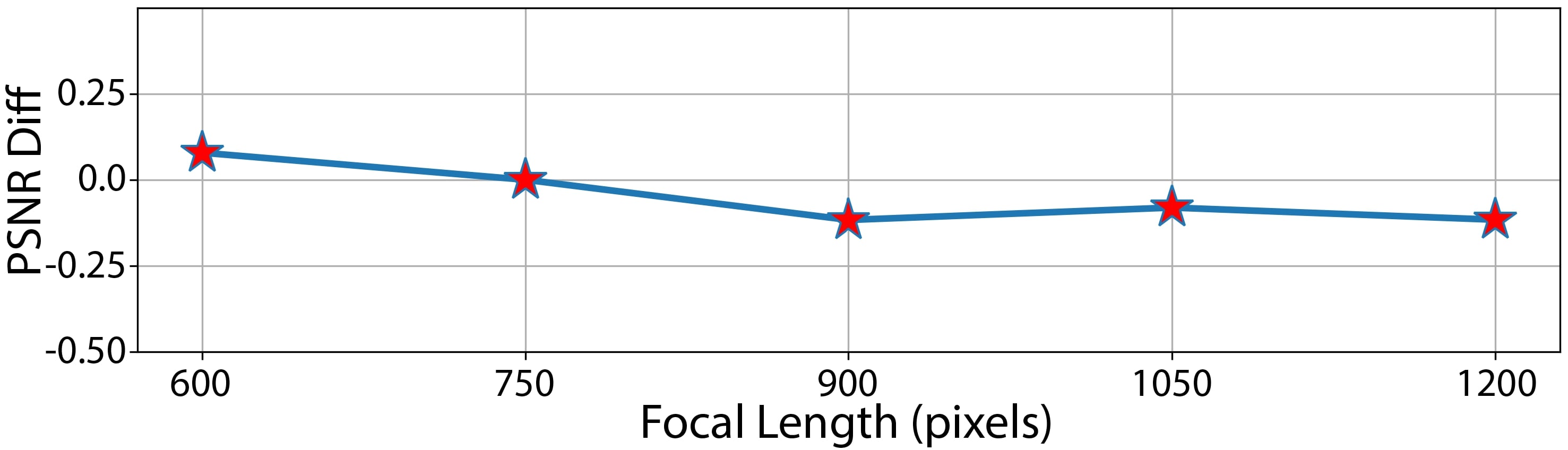}
\vspace{-2mm}
\caption{\small Visual plots showing the performance of GeoFill with different focal lengths. PSNR diff is computed by using GeoFill with new focal length subtract GeoFill with focal length equals to 750. }
\label{fig: focal_study}
\vspace{-2mm}
\end{figure}

\section{Performance w.r.t Intrinsic Parameters}
\label{ssec:focal_study}
GeoFill handles incoming image pairs using fixed camera intrinsic parameters instead of explicitly knowing the ground-truth camera intrinsic parameters. 
In the main paper, we use fixed camera intrinsic parameters by setting the focal length of all images to 750 pixels and the principal point to the center of the image.
It is intuitive to set the principal point to the center of the images with unknown intrinsic parameters, therefore, we focus on studying the effect of focal lengths. 
We compare the performance of GeoFill with the camera focal lengths of 600, 750, 900, 1050, and 1200 pixels. As shown in Fig.~\ref{fig: focal_study}, GeoFill with different focal lengths has very slight differences in terms of PSNR.
There is a slight trend that the performance drops as the focal length increases. We believe this indicates that the ground-truth focal length is close to 600 and higher focal lengths make the optimization have a harder time finding improved relative poses.
Nevertheless, GeoFill can still adapt to different focal lengths by jointly optimizing depth scale, offset, and relative pose, therefore, it still can render similar images across a variety of focal lengths. 

\begin{figure*}[t]
\centering
\includegraphics[width=0.99\linewidth]{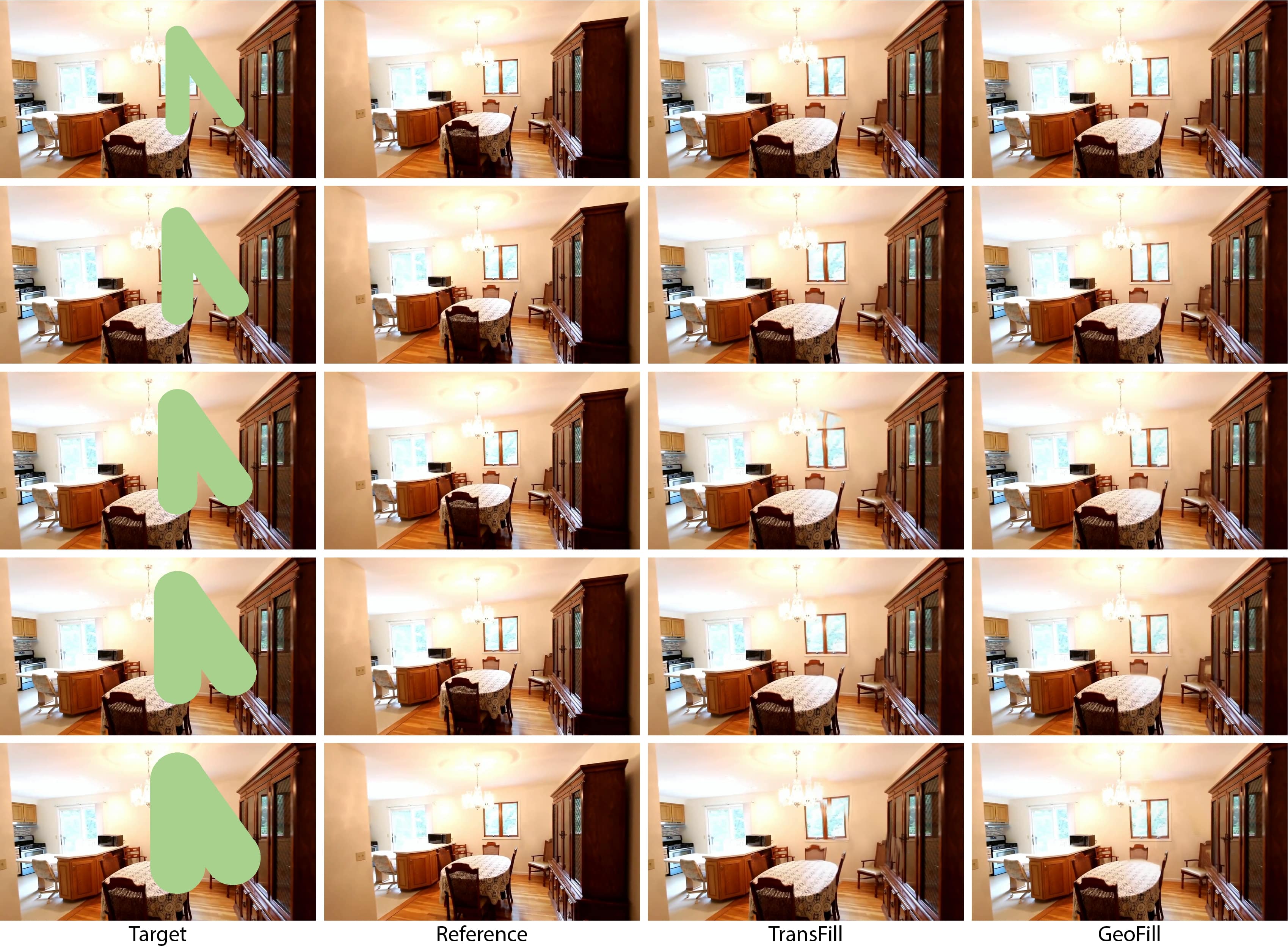}
\vspace{-2mm}
\caption{\small Qualitative comparisons of our approach against TransFill with different hole sizes. \textbf{Please zoom in} to see that ours looks good but there are broken structures, ghosting, and distortion artifacts in TransFill.}
\label{fig: hole_abl_vis}
\end{figure*}

\begin{table}[t]\setlength{\tabcolsep}{13pt}
\setlength{\abovecaptionskip}{0pt}
\centering
\footnotesize
\caption{Initial alignment comparisons of our method compared to TransFill without the CST Module on MannequinChallenge dataset (FD=10).}
\begin{tabular}{|c|c|c|c|}
\hline
Model &PSNR$\uparrow$&SSIM$\uparrow$&LPIPS$\downarrow$\\ \hline\hline
GeoFill & \bf{28.85} & \bf{0.9702} & \bf{0.0553} \\\hline
GeoFill (no CST) & 26.84 & 0.9626 & 0.0615 \\\hline
TransFill & 28.01 & 0.9680 & 0.0569 \\\hline
TransFill (no CST) & 24.04 & 0.9526 & 0.0760 \\
\hline
\end{tabular}
\label{tab:no_cst_MannequinChallenge}
\vspace{-2mm}
\end{table}

\section{Further Analyses of Initial Alignments}
\label{ssec:no cst}
\begin{figure*}[t]
\centering
\includegraphics[width=0.99\linewidth]{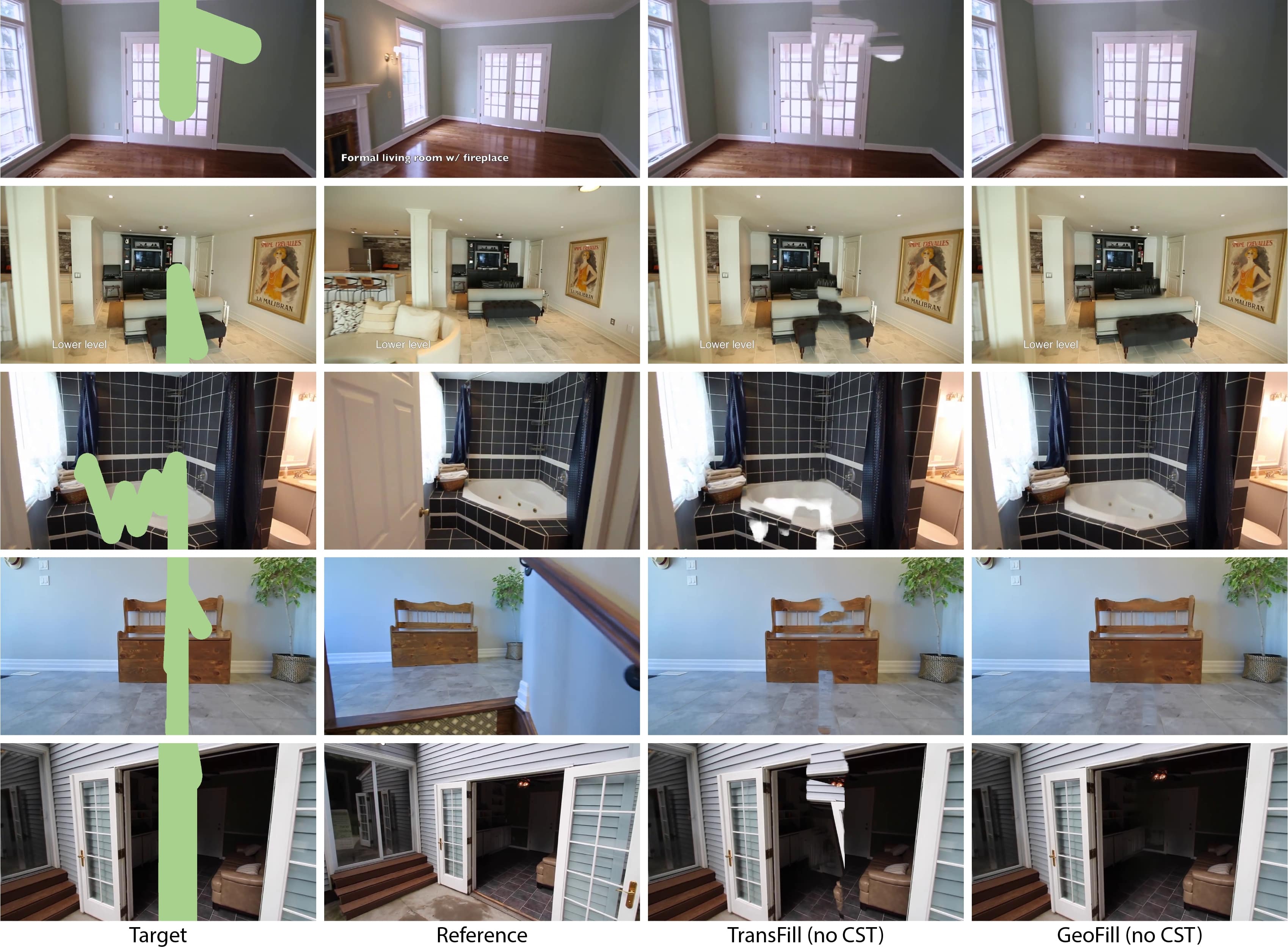}
\vspace{-2mm}
\caption{\small Visual comparisons of GeoFill and TransFill without the CST module.}
\vspace{-4mm}
\label{fig: no_cst}
\end{figure*}

GeoFill adopts the CST module from TransFill to further improve any small residual spatial misalignments and correct color and exposure differences. We first visually compare the quality of our single proposal to the merged proposal from TransFill without the CST on RealEstate10K dataset. As shown in Fig.~\ref{fig: no_cst}, the single proposal from GeoFill is significantly more accurate than merged heuristic proposals from TransFill, demonstrating the superiority of our approach over TransFill. 
Additionally, we also show the quantitative comparisons of GeoFill and TransFill without the CST module on the MannequinChallenge dataset. As shown in Table~\ref{tab:no_cst_MannequinChallenge}, we find GeoFill without the CST module has a huge advantage over TransFill merged homographies.

\section{Visualizations for Hole Ablation Study}
\label{ssec:hole ablation vis}
In the main paper, we show the quantitative comparisons between our approach and TransFill with various hole sizes. Here, we provide visual comparisons to better understand the performance boost for larger holes. 
We simulate larger holes by generating the same hole shape with larger stroke width. As shown in Fig.~\ref{fig: hole_abl_vis}, GeoFill has a robust performance while TransFill has ghosting artifacts and misalignments as the hole grows larger.

\begin{figure*}[t]
\centering
\includegraphics[width=0.99\linewidth]{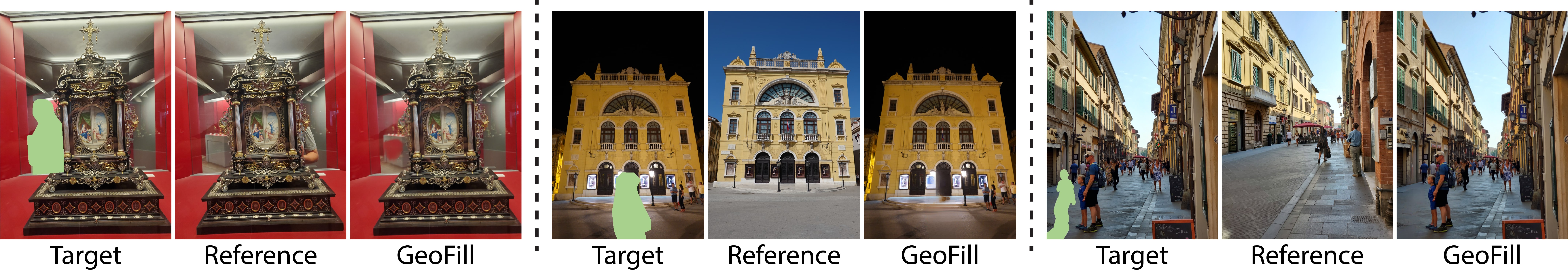}
\vspace{-2mm}
\caption{\small Visual examples of failure cases of GeoFill.}
\vspace{-3mm}
\label{fig: failure_cases}
\end{figure*}

\begin{table}[t]\setlength{\tabcolsep}{6pt}
\setlength{\abovecaptionskip}{0pt}
\centering
\footnotesize
\caption{User study results of GeoFill against ProFill, OPN, and TransFill.}
\begin{tabular}{|c|c|c|c|c|}
\hline
& \multicolumn{2}{c|}{Filtered} & \multicolumn{2}{c|}{Non-Filtered}\\ \hline
Model & PR & p-value  & PR & p-value \\ \hline\hline
ProFill & 100\% & $p < 10^{-6}$ & 96.25\% & $p < 10^{-6}$ \\\hline
OPN & 97.37\% & $p < 10^{-6}$ & 95.00\% & $p < 10^{-6}$ \\\hline
TransFill & 70.90\% & $p < 2\times10^{-3}$ & 68.13\% & $p < 2\times10^{-3}$ \\\hline
\end{tabular}
\vspace{-4mm}
\label{tab:user_study}
\end{table}

\section{Average Running Time of GeoFill}
\label{ssec:timing}
We randomly sampled 50 images at 1280x720 pixels and compute the average time of each step. Monocular depth estimation takes 3.83s, sparse correspondence estimation takes 0.596s, triangulation takes 0.0009s, initial relative pose takes 0.0052s, joint optimization takes 58.2s, mesh rendering takes 1.03s, refinement and merging step takes 2.53s.
The reported time uses default parameters described in the experiment section.
Although the joint optimization step takes up the vast majority of the time, its current implementation is naive and not optimized. If desired, 
various engineering optimizations could be made such as using a custom kernel with proper low-level optimizations such as fusion for the renderer instead of a naive pure PyTorch implementation, using FP16 mode, using only the sparse edge map pixels during optimization (these are quite sparse so significant acceleration should be possible), using second-order optimization techniques that could potentially converge in fewer steps, carefully tuning input resolution, number of pyramid levels, iteration limits, break thresholds, etc. We considered these to be lower-level engineering details that we did not focus on in our paper's implementation, since we were focusing more on research aspects.

\section{User Study}
\label{ssec:user_study}
To better evaluate the performance of GeoFill against other baselines, we conduct a user study via Amazon Mechanical Turk (AMT).
We compare our method against OPN, ProFill, and TransFill by showing users image pairs with binary choice questions. 
The users are requested to choose the inpainting results that look more realistic and faithful. 
To improve the quality of collected data, we adopt a qualification test with trivial questions to filter noisy results.
For each method pair, we randomly sampled 80 examples in RealEstate10K dataset with FD=50, and each example was evaluated by 7 independent users. 
We present two approaches to computing the preference rate. The first one is the filtered approach, in which we filter the responses to retain only those where one method is ``preferred" if 6 or more users select it. The filtering helps suppress noise in the responses of Mechanical Turk workers, whose work quality can vary. The second one is the non-filtered approach where we retain all responses and choose the method as ``preferred" where a simple majority of 4 or more users select it. 
We reported GeoFill’s Preference Rate (PR) in Table~\ref{tab:user_study}. 
GeoFill has much higher preference rates against OPN and ProFill. 
Compared against TransFill, we receive a PR around 70\% on filtered and non-filtered approaches. TransFill is still very robust on small holes and relatively small camera motions in the randomly sampled data. Therefore, GeoFill is favored by users over TransFill but less strongly than in the other comparisons. We performed a one sample permutation t test with $10^6$ simulations using the null hypothesis that each pair are preferred equally by users: the p-values are all sufficiently small that the preference for our method is statistically significant.

\begin{figure}[t]
\centering
\includegraphics[width=0.99\linewidth]{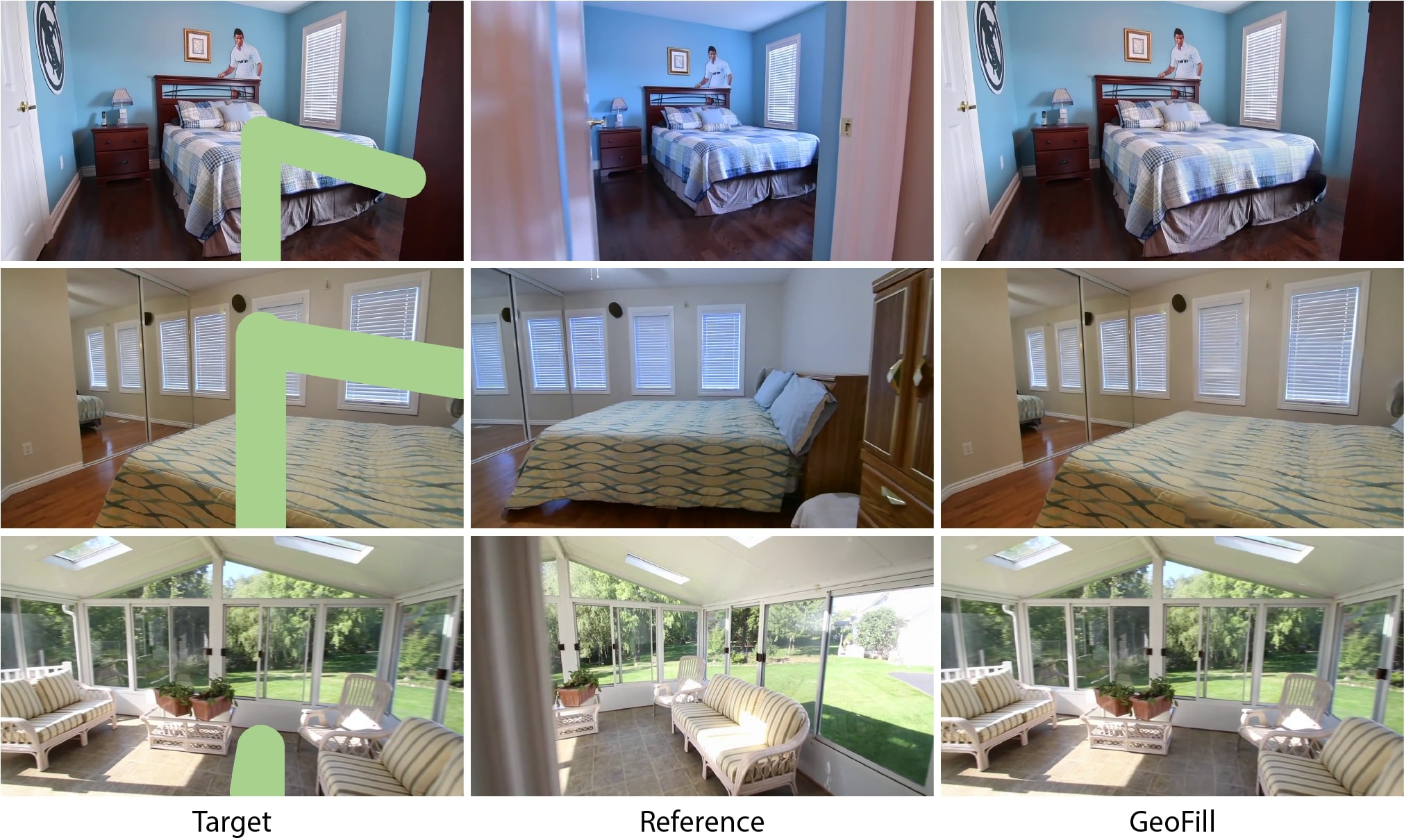}
\vspace{-2mm}
\caption{\small Qualitative results of GeoFill handling some common appearance changes such as in white balance and exposure due to camera movement.}
\vspace{-3mm}
\label{fig: appear_change}
\end{figure}

\vspace{-2mm}
\section{Handling Appearance Changes from Camera Movement}
\label{ssec:appear}
As we stated in the main paper, we focus on the common scenario of capturing photos with the same camera {\em freely} moving around. However, there are potential appearance changes of the same parts of the scene due to the camera movement, for example, changes due to automatic exposure or automatic white balance between source and target images. This is a common yet non-trivial challenge when applying GeoFill in real-world applications. In this section, we show some visual examples of image pairs with appearance changes in the dataset. As shown in Figure~\ref{fig: appear_change}, GeoFill still inpaints plausible results even when the appearance of the same part of the scene is different between source and target images.

\section{Failure Cases}
\label{ssec:failure}
We show some failure cases of GeoFill under extreme conditions. Fig.~\ref{fig: failure_cases} shows three common failure cases of GeoFill. The image pair on the left contains transparent surfaces in the images. These objects often cause monocular depth estimators to fail and can lead to bad optimization results. In the second failure case, the drastic changes in the lighting environment affect the feature correspondence matching and depth prediction, which makes the final result from GeoFill less accurate. In the last case, dynamic objects, e.g., pedestrians, make our optimization module estimate inaccurate parameters. We discuss in the last section of our main paper ways that future work might address these issues.

\section*{Additional Visual Results}
\begin{figure*}[t]
\centering
\includegraphics[width=0.99\textwidth]{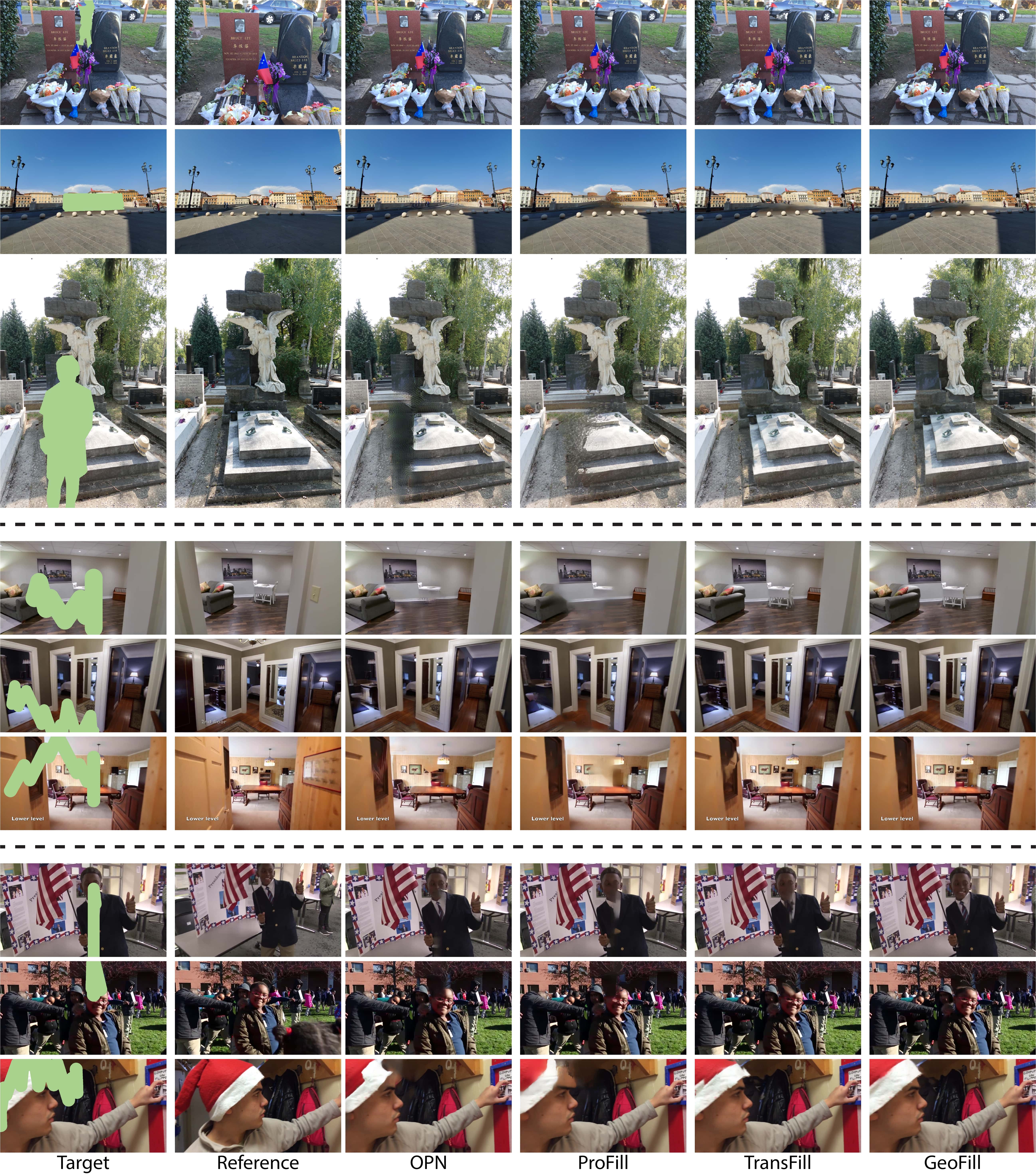}
\caption{Qualitatively comparison of GeoFill against other baselines on user-provided images (top 3 rows), RealEstate10K (mid 3 rows), and MannequinChallenge dataset (last 3 rows). 
}
\label{fig: suppl_vis1}
\end{figure*}

\begin{figure*}[t]
\centering
\includegraphics[width=0.99\textwidth]{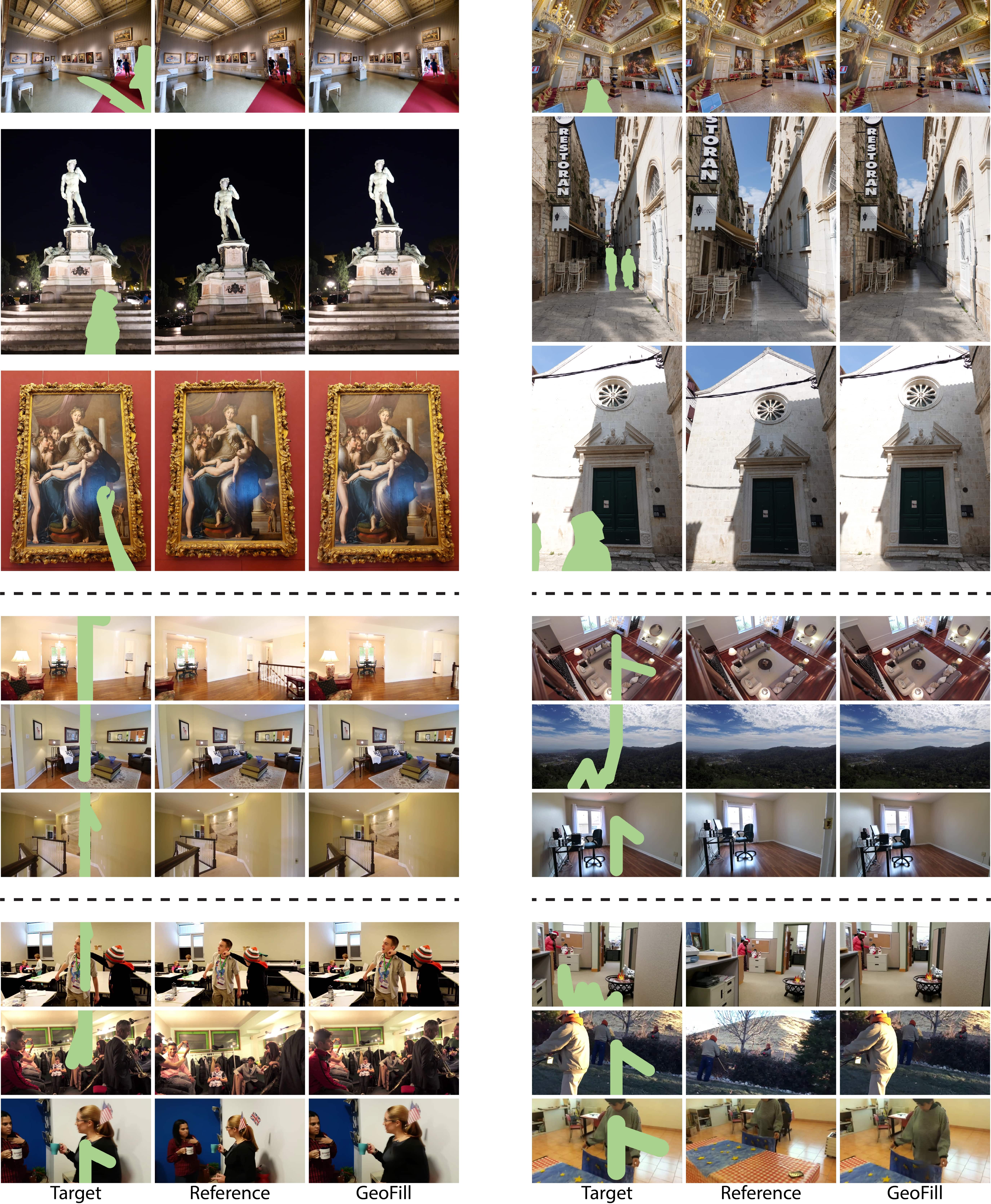}
\caption{Visual illustration of inpaiting performance of GeoFill on user-provided images, RealEstate10K, and MannequinChallenge dataset.
}
\label{fig: suppl_vis2}
\end{figure*}

We include additional qualitative comparisons of GeoFill against other baselines in Fig.~\ref{fig: suppl_vis1}. Additionally, we also show the inpainting performance of GeoFill on user-provided images, RealEstate10K, and MannequinChallenge dataset in Fig.~\ref{fig: suppl_vis2}.

\newpage

\newpage

\end{document}